\documentclass[conference]{IEEEtran}
\IEEEoverridecommandlockouts
\usepackage{cite}
\usepackage{amsmath,amssymb,amsfonts}
\usepackage{textcomp}
\usepackage{hyperref}
\hypersetup{
    colorlinks=true,
    linkcolor=black,
    urlcolor=black,
    citecolor=black
}
\usepackage[table]{xcolor}
\usepackage{bbm}
\usepackage{booktabs}
\usepackage{makecell}
\usepackage{multirow}
\usepackage{graphicx}
\usepackage{algorithm}
\usepackage{algpseudocode}
\usepackage{subcaption}
\usepackage[font=small]{caption}

\def\BibTeX{{\rm B\kern-.05em{\sc i\kern-.025em b}\kern-.08em
    T\kern-.1667em\lower.7ex\hbox{E}\kern-.125emX}}
\begin{document}

\title{SoGuDiff: Socially Guided Diffusion for Steerable, Norm-Grounded Robot Navigation
\thanks{$^1$Department of Electrical and Computer
Engineering, University of Waterloo, Waterloo, Canada (e-mails: \{cschaible, yash.pant, stephen.smith\}@uwaterloo.ca).
\protect\\ \indent $^2$Department of Computing and Software, McMaster University, Hamilton, Canada (e-mail: jih21@mcmaster.ca). This work was done while Haoran Ji was an undergraduate research fellow at the University of Waterloo.
\protect\\ \indent This research is supported in part by the Natural Sciences and Engineering Research Council of Canada (NSERC). Resources used in preparing this research were provided, in part, by the Province of Ontario, the Government of Canada through CIFAR, and companies sponsoring the Vector Institute.
\protect\\ \indent \textbf{Code:} \url{https://github.com/schaiblc/SoGuDiff}.}
}

\author{Christian Schaible$^1$, Haoran Ji$^2$, Yash Vardhan Pant$^1$, Stephen L. Smith$^1$}

\maketitle

\begin{abstract}
Beyond collision avoidance, socially competent robot navigation requires adherence to implicit social conventions that vary across contexts, cultures, and deployment requirements. Many conventional navigation policies learn a single normative behavior, either through reinforcement learning against a fixed reward function or imitation of human demonstrations, exposing no interface for adjusting that conduct at runtime. We present a diffusion-based navigation framework whose social behavior can be tuned at deployment: a desired style is specified, such as how closely the robot passes, which side it yields to, or how much it defers to groups, and the planner adapts accordingly. Continuous style axes can be followed independently or composed, spanning a behavioral space rather than discrete, primitive-based specifications. A feasibility projection layer separates learned social behavior from kinematic feasibility and collision avoidance. A single-axis sweep illustrates a tradeoff curve that strictly dominates the evaluated fixed-behavior baseline configurations, and stylistic differences are replicated in real-world demonstrations.
\end{abstract}

\section{Introduction}

The conventions that govern movement among pedestrians are not universal. They depend on culture, environment, and task. A level of caution suited to a quiet care facility becomes obstructive in a busy concourse, and passing conventions differ between regions with left- and right-hand traffic. A robot operating across such settings therefore cannot rely on a single fixed behavior, and instead needs the ability to configure its social conduct to context. Users have been shown to prefer explicit, predictable conventions from robots, where adherence measurably improves perceived safety~\cite{gallo2023investigating, UserStudy2}.

Most existing methods commit to one behavior. Model-based planners encode norms
as hand-tuned potential fields or constraints~\cite{SFM,
ORCA}, which are interpretable but require manual
re-tuning to change. Reinforcement learning (RL) policies for crowd navigation~\cite{CADRL, CrowdNavSARL, DSRNN} embed a single normative behavior in their weights, so altering it means
redesigning the reward and retraining. Multi-objective RL trades objectives off at runtime via a preference vector~\cite{deheuvel2025, cheng2023morl}, but such policies are unimodal and select one behavior where several are valid. Compositional diffusion offers a different route. ComposableNav composes navigation primitives
at deployment~\cite{ComposableNav}, but from discrete binary
specifications realized by separately trained models, each expressing a targeted behavior rather than a graded one.

\begin{figure}[t]
    \centering
    \includegraphics[width=1.0\linewidth]{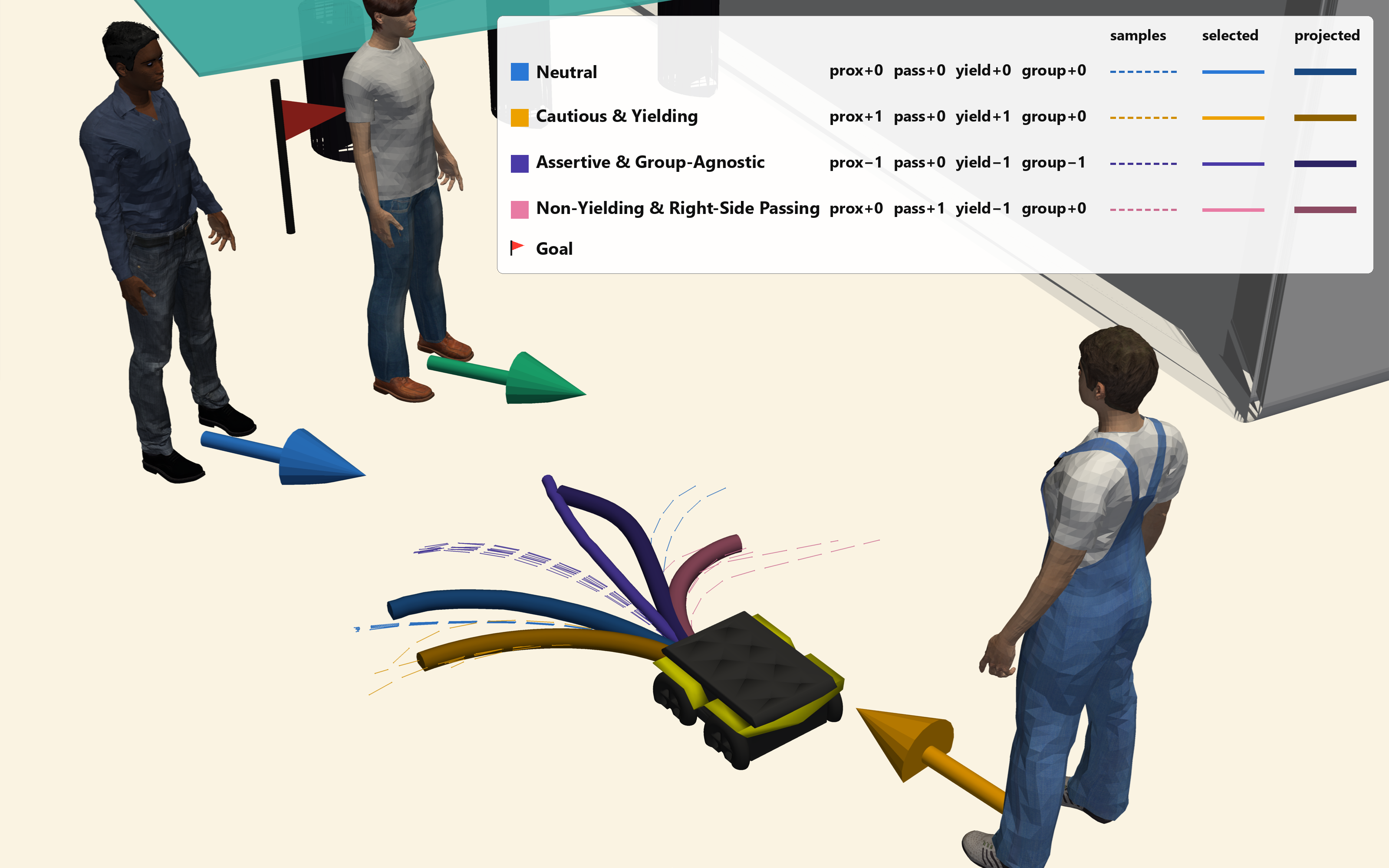}
    \caption{SoGuDiff steers a single diffusion planner along interpretable, composable social style axes at deployment. For each style, candidate trajectories are sampled, one is selected and feasibility projection ensues. Varying the style vector, without retraining, yields controllable, compositional behaviors. Human meshes are from HuNavSim \cite{HuNavSim}.}
	\label{teaser}
\end{figure}

We instead treat social navigation as a continuously parameterized family of
behaviors generated by one model. Social behavior is inherently multimodal, since passing a group on
either side can be equally valid, a structure that single-mode planners collapse
but diffusion models represent naturally~\cite{janner2022diffuser,
chi2023diffusionpolicy}. We define four
interpretable style axes (proxemic conservatism, passing-side convention,
yielding disposition and group deference), drawn from social-navigation conventions~\cite{moussaid2010groups, katyal2020groupaware,
johnson2018passing}. Using demonstrations that each only carry a label on one axis, we train a single cross-attention conditional diffusion
model by imitation. Per-axis classifier-free guidance (CFG) with structured conditioning dropout~\cite{CFG, DecisionDiffuser} then steers each axis independently and composes them at inference, producing styles the model was never shown jointly. A feasibility projection layer refines diffused trajectories towards ones that respect robot dynamics and obstacle clearances. Fig.~\ref{teaser} illustrates the resulting control space.

This paper introduces the approach, \textbf{So}cially \textbf{Gu}ided \textbf{Diff}usion (SoGuDiff), making the following contributions:
\begin{itemize}
  \item We frame social navigation as a steerable, continuously
  parameterized family of behaviors realized by a single conditional diffusion planner, with four interpretable style axes specified at deployment rather than fixed in training.
  \item We employ a per-axis CFG scheme that steers style axes independently and composes them at inference, from a model trained only on single-axis-labeled demonstrations.
  \item We conduct evaluations showing that: single-axis and composed styles produce interpretable behaviors; sweeping an axis yields an efficiency-vs-sociality tradeoff curve that strictly dominates fixed-behavior baseline planners; and styled behavior transfers to real-world deployment.
\end{itemize}

\section{Related Work}

\subsection{Social Robot Navigation and Adjustable Behavior}

Early methods model social behavior with reactive rules, such as the repulsive
forces of the social force model~\cite{SFM} or the velocity
constraints of reciprocal collision avoidance~\cite{ORCA}. Learning-based crowd
navigation instead trains policies by reinforcement learning, from value-based
collision avoidance~\cite{CADRL} to attention- and graph-based interaction
models~\cite{CrowdNavSARL, DSRNN, RGL, HEIGHT} and
transformer policies with preference learning~\cite{NaviSTAR}.
Optimization-based planners such as SICNav couple prediction and planning in a
bilevel program with explicit safety constraints~\cite{SICNav}. In each case, the social behavior is fixed once the
reward is designed or the cost is tuned.

Multi-objective RL is the established exception, conditioning a policy on a preference vector over reward components in general~\cite{deheuvel2025} and crowd~\cite{cheng2023morl} navigation. Such policies are unimodal and select an action one step at a time. In contrast, we employ a generative planner that produces receding-horizon trajectories from the full multimodal distribution and separates the strength of a style from its requested value.

\subsection{Social Conventions and Behavior Specification}

Proxemics, the regulation of personal space, is the most widely modeled social
norm~\cite{mavrogiannis2023core, singamaneni2024survey}, following Hall's
account of interpersonal zones~\cite{hallproxemics}, yet social competence extends beyond it. Surveys identify passing on a consistent side, resolving
head-on encounters, respecting groups, and moving legibly as further conventions
that shape comfort~\cite{singamaneni2024survey}. Pedestrians tend to avoid walking
through social groups~\cite{moussaid2010groups}, group-aware policies that preserve
formations reduce discomfort at a small cost in
efficiency~\cite{katyal2020groupaware}, and passing-side conventions have been
learned from pedestrian observation~\cite{johnson2018passing}. Encoding such
norms as weighted terms in a navigation cost is long-established practice~\cite{kirby2009companion, mavrogiannis2023core}, and the weights may be set by hand or recovered from human demonstration by inverse reinforcement learning~\cite{kretzschmar2016irl}. A more recent alternative scores candidate actions against a vision-language model's response to a hand-written prompt template, which can be manually reworded for different conventions~\cite{VLM}. We instead ground our four style axes in these conventions and, rather than fixing one weighting or rewriting a prompt, use a cost of
this kind only as a means of producing continuous style-labeled demonstrations.

\subsection{Diffusion Planning and Controllable Generation}

Diffusion models are effective trajectory and action generators for
planning~\cite{janner2022diffuser, chi2023diffusionpolicy, LDP}. Conditions can be combined at sampling time by composing score estimates, generalizing CFG with either separately trained models or a single model queried under different conditions~\cite{liu2022compositional}. The Decision Diffuser applies this to compose constraints and skills that were never observed
together~\cite{DecisionDiffuser}. In autonomous driving, controllable traffic
generation steers a diffusion model at test time with differentiable rule
gradients~\cite{CTG}. Legibility Diffuser guides a goal-conditioned diffusion policy toward intent-expressive trajectories while decaying the guidance weight along the trajectory to preserve task success~\cite{LegibilityDiffuser}. Within social navigation, diffusion has served as an action representation inside reinforcement learning, made controllable through test-time guidance toward task variants such as static obstacle avoidance and person following~\cite{COLSON}. Elsewhere, diffusion has been used for pedestrian prediction feeding a downstream controller~\cite{SICNavDiff}.

The closest work is ComposableNav~\cite{ComposableNav}, which composes
navigation primitives by summing the noise predictions of separately trained
models, each obtained through supervised pretraining followed by RL
fine-tuning; the same principle generalizes trajectory planning to unseen scenes
by combining separately trained per-scene experts~\cite{RSTP}. The ComposableNav specifications are discrete, binary instructions, whereas ours are continuous, signed values denoting
degrees of a convention. ComposableNav trains one model per primitive, whereas we train one
model and separate the axes through conditioning, avoiding per-primitive training altogether. Our contribution lies in the control interface: a single planner
whose conduct is set by a continuous style vector, composed at inference from
supervision that only ever varied one axis at a time.

\section{Problem Formulation}
\label{sec:prob_form}

\begin{figure*}[t]
    \centering
    \includegraphics[width=1.0\textwidth]{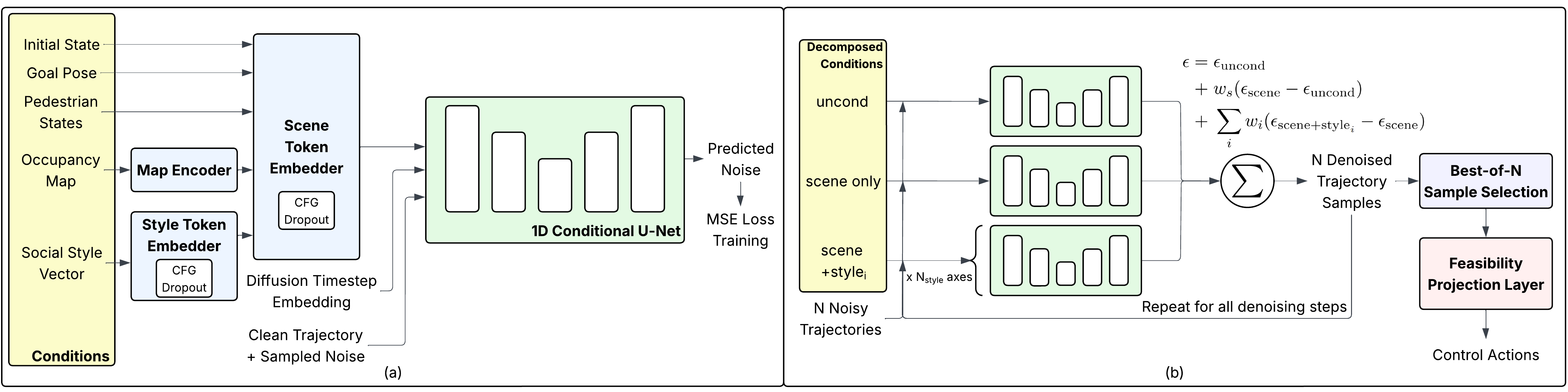}
    \caption{Overview of SoGuDiff. (a) Training: scene inputs (initial state, goal, pedestrian states, and occupancy map) and the social style vector are embedded as tokens, with structured conditioning dropout over the scene and style tokens. A one-dimensional conditional U-Net is trained to denoise trajectories. (b) Inference: the network is queried under decomposed conditions (unconditional, scene, and scene-plus-each-style-axis) and combined by per-axis classifier-free guidance to steer and compose style. Candidate trajectories are denoised in parallel, the best is selected, and feasibility projection either produces feasible actions or triggers a controlled stop if acceptance criteria are not met.}
	\label{SocialStyleUNet}
\end{figure*}

We consider a robot navigating toward a goal among pedestrians and static
obstacles. The robot's state at time $t$ is
$x_t=(\mathbf{p}^r_t,\theta_t,v_t,\omega_t)$, comprising its position
$\mathbf{p}^r_t\in\mathbb{R}^2$, heading $\theta_t$, and linear and angular
velocities $v_t,\omega_t$. The state evolves under a kinematic model $x_{t+1}=f(x_t,u_t)$, driven by controls $u_t=(a_t,\alpha_t)$, linear and angular acceleration.
The robot is tasked with reaching a goal position $\mathbf{g}\in\mathbb{R}^2$
while avoiding collision with up to $P$ nearby pedestrians, each described by position and velocity vectors $(\mathbf{p}^{j},\mathbf{v}^{j})$, and with static
obstacles given by a local occupancy map $\mathcal{M}\in\{0,1\}^{H\times W}$. Together, the robot's current
state, goal, pedestrian states, and occupancy map form the scene context
$c=\{x_0,\mathbf{g},\mathcal{P},\mathcal{M}\}$, where
$\mathcal{P}=\{(\mathbf{p}^{j},\mathbf{v}^{j})\}_{j=1}^{P}$.

A plan is a trajectory $\tau=(\mathbf{p}^r_1,\dots,\mathbf{p}^r_T)$, a sequence
of $T$ positions at a fixed time step $\Delta t$, executed in receding-horizon fashion.
A trajectory is admissible only if
some control sequence $u_{0:T-1}$ drives the state through $f(x_t,u_t)$ from $x_0$, tracking $\tau$, respects
the robot's kinematic limits, and keeps it clear of pedestrians and static obstacles at every step.

Beyond reaching the goal safely, the robot should conform to the social
conventions of its deployment, which vary with context.
We make the desired convention an explicit input: a social style vector
$\mathbf{s}=[s_{\mathrm{prox}},s_{\mathrm{pass}},s_{\mathrm{yield}},s_{\mathrm{group}}]
\in[-1,1]^4$, whose four axes denote proxemic conservatism, passing-side
convention, yielding disposition, and group deference. For the proxemic, yielding and group axes, $s_i=0$ is the default neutral convention, raising $s_i>0$ produces increasingly deferential conduct, and decreasing $s_i<0$ relaxes convention adherence in favor of
efficiency. For the passing axis, $\mathrm{sign}(s_{\mathrm{pass}})$ selects the side and $|s_{\mathrm{pass}}|$ is the degree to which it is enforced, so $s_{\mathrm{pass}}=0$ denotes no side preference.

\noindent\textbf{Problem statement:} Given the scene context $c$ and a desired social style $\mathbf{s}$, generate an admissible trajectory $\tau$ toward the goal $\mathbf{g}$ that conforms to $\mathbf{s}$.

\section{Methodology}

We present the three major components of SoGuDiff: a
style-conditioned diffusion model (Sec.~\ref{sec:arch}), a compositional guidance
scheme (Sec.~\ref{sec:guidance}),
and a feasibility projection layer (Sec.~\ref{sec:projection}). The single-axis-labeled training demonstrations are described in
Sec.~\ref{sec:data}. The pipeline is shown in Fig.~\ref{SocialStyleUNet}.

\subsection{Style-Conditioned Diffusion Model}
\label{sec:arch}

We model the distribution over trajectories with a conditional denoising diffusion
probabilistic model (DDPM)~\cite{ho2020denoising}. During training, a forward process
gradually corrupts a clean trajectory $\tau^0$ into Gaussian noise over $K$ steps,
\mbox{$q(\tau^{k}\mid\tau^{k-1})$}$=\mathcal{N}(\tau^k;\sqrt{1-\beta_k}\,\tau^{k-1},\beta_k\mathbf{I})$,
and the network $\epsilon_\theta$ learns to reverse it by predicting the noise
added at each step. Conditioning on scene context and style vector, the objective is
\begin{equation}
  \mathcal{L}=\mathbb{E}_{k,\tau^0,\epsilon}
  \big[\,\lVert\epsilon-\epsilon_\theta(\tau^{k},k,c,\mathbf{s})\rVert^2\,\big],
  \quad \epsilon\sim\mathcal{N}(\mathbf{0},\mathbf{I}).
  \label{eq:objective}
\end{equation}
At inference, trajectories are denoised with the denoising
diffusion implicit model (DDIM) sampler~\cite{DDIM} for efficiency.

The network conditions on $c$ and $\mathbf{s}$ through a token-based transformer encoder feeding a one-dimensional conditional U-Net (Fig.~\ref{SocialStyleUNet}a). The initial state, goal, and pedestrian states (expressed in the robot-centric frame) are mapped to tokens, the occupancy map is compressed into tokens by a convolutional map encoder, and each style axis is embedded as its own token. Tokens for absent pedestrians (when fewer than $P$ are detected) are excluded by a
validity mask, propagated as an attention mask. Scene and style tokens are concatenated and processed jointly by a self-attention encoder, so that style and scene interact before conditioning the denoiser. The U-Net then generates the trajectory through residual blocks that cross-attend to the encoded tokens, while the diffusion step enters as an embedding that modulates the residual block features.  The conditioning is a compact set of explicit state and style tokens rather than a high-dimensional perceptual or language representation, keeping the model lightweight.

During training, the conditioning mask is drawn from a structured categorical distribution over exactly the conditioning patterns queried at inference: the unconditional mask, the scene-only mask, and the four masks retaining the scene and a single style axis, with the remainder fully conditioned. Training every mask in one model enables joint and per-axis schemes (Sec.~\ref{sec:guidance}) to be contrasted (Sec.~\ref{sec:ablation}).

\subsection{Compositional Guidance and Sampling}
\label{sec:guidance}

Structured dropout during training lets the same network be queried under
different conditioning subsets at inference, which we exploit for compositional control (Fig.~\ref{SocialStyleUNet}b). Writing $\varnothing$ for a token group replaced by its learned null embedding and $\mathbf{s}^{(i)}$ for retaining only non-null style axis $i$ atop the scene, the guided noise estimate, inspired by \cite{CFG, DecisionDiffuser}, is
\begin{equation}
\begin{aligned}
  \hat{\epsilon}=\;&\epsilon_\theta(\tau^{k},k,\varnothing,\varnothing)
  + w_s\big[\epsilon_\theta(\tau^{k},k,c,\varnothing)-\epsilon_\theta(\tau^{k},k,\varnothing,\varnothing)\big]\\
  &+\sum_{i}w_i\big[\epsilon_\theta(\tau^{k},k,c,\mathbf{s}^{(i)})-\epsilon_\theta(\tau^{k},k,c,\varnothing)\big].
\end{aligned}
  \label{eq:cfg}
\end{equation}

Here, $w_s$ sets the strength of scene conditioning and each $w_i$ independently sets the strength of its axis. Following the two-scale form used in InstructPix2Pix \cite{InstructPix2Pix}, each style delta is taken relative to the scene-conditional estimate $\epsilon_\theta(\tau^{k},k,c,\varnothing)$ rather than the unconditional one. Hence, a style acts as a graded adjustment on top of the scene rather than re-specifying it. The total guidance weight $\sum_i w_i$ is held constant and split equally over the active non-zero axes, so that a composed style is not guided more strongly in aggregate than a single-axis one.

Setting $w_i=0$ removes an axis' contribution and raising it
strengthens that behavior, so axes are adjusted and combined without retraining. Since $\mathbf{s}^{(i)}$ and $w_i$ enter separately, the desired style and the strength with which it is imposed remain independent controls. The alternative is to retain all style tokens in a single conditional query and apply one delta with one weight. This evaluates the joint conditional directly but offers no per-axis control (Sec.~\ref{sec:ablation} compares the two). All conditioning variants
are evaluated in a single batched forward pass, so guidance adds no sequential denoising steps.

The denoiser produces $N$ candidate trajectories, also batched in parallel. Rather than re-ranking by the social cost, which the sampled candidates already consider, a selector evaluates a simpler goal-directed objective. This cost combines collision penalties, a goal-progress reward and a small smoothness regularizer. The lowest-cost candidate $\tau^{\star}$ is returned.

\subsection{Feasibility Projection}
\label{sec:projection}

The selected trajectory captures social intent but carries no kinematic or safety guarantee, since a diffusion model cannot ensure either on its own. A projection layer refines $\tau^{\star}$ toward a dynamically feasible, collision-avoiding trajectory by solving a soft-constrained optimal control problem (OCP) with \textit{acados}~\cite{acados}, tracking $\tau^{\star}$ as a reference under given dynamics, actuation limits, and obstacle clearance constraints. Using a unicycle model for $f$, the OCP is
\begin{equation}
\begin{aligned}
  \min_{x_{0:T},\,u_{0:T-1},\,\xi\ge0}\;&\sum_{t=1}^{T}\lVert\mathbf{p}^{r}_t-\mathbf{p}^{\star}_t\rVert_Q^2
  +\sum_{t=0}^{T-1}\lVert u_t\rVert_R^2\\
  &+\sum_{t=1}^{T}\sum_j\big(\rho_1\,\xi^j_{t}+\rho_2\,(\xi^j_{t})^2\big)\\
  \text{s.t.}\;\;&
  \begin{array}[t]{@{}l@{}}
  \begin{array}[t]{@{}ll@{}}
    x_{t+1}=f(x_t,u_t),\quad &x_0=(\mathbf{p}^{r}_0,\theta_0,v_0,\omega_0),\\[3pt]
    
      0\le v_t\le v_{\max}, & \lvert\omega_t\rvert\le\omega_{\max},\\[2pt]
      \lvert a_t\rvert\le a_{\max}, & \lvert\alpha_t\rvert\le\alpha_{\max},
    \end{array}\\[3pt]
    \lVert\mathbf{p}^{r}_t-\mathbf{o}^j_{t}\rVert^2\ge(r^r+r^j+r^s)^2-\xi^j_{t},
  \end{array}
\end{aligned}
  \label{eq:ocp}
\end{equation}
where $\mathbf{p}^{\star}_t$ are the waypoints of the selected trajectory, tracked by the cost with weight $Q$. Control effort is regularized with the weight $R$, and the slack $\xi^j_{t}\ge0$ is introduced on the clearance constraint for obstacle $j$ at step $t$ (inflated by safety radius $r^s$). Each obstacle $j$ has radius $r^j$ and center $\mathbf{o}^j_{t}=\mathbf{o}^j_{0}+t\Delta t\,\mathbf{v}^j$, propagated at constant velocity across the $T$-step, $\Delta t$-spaced horizon.  Pedestrians enter as moving obstacles, while static obstacles
sampled from the occupancy map enter with
$\mathbf{v}^j=\mathbf{0}$. The slack penalty weights ($\rho_1,\rho_2$) discourage constraint violation without hard-bounding it, keeping the problem solvable from infeasible warm starts.

The OCP is solved by sequential quadratic programming (SQP) to a fixed iteration budget per control cycle. A solution is accepted only if the largest slack over all obstacles and steps, and the deviation from $\tau^{\star}$, both fall below fixed thresholds, checked after the solve rather than enforced within it. The slack threshold preserves some clearance inflation, so an accepted trajectory remains collision-free with respect to the physical radii. Otherwise, a controlled stop is commanded.

This reflects a separation of concerns: the diffusion model owns social behavior, style, and multimodality, while the OCP accounts for kinematic feasibility and collision avoidance. The kinematic model is interchangeable, so the framework should extend to other dynamics by substituting $f$. Unlike methods that embed control barrier or Lyapunov constraints inside the denoising process~\cite{SafeDiffuser, COBL}, the projection acts independently after generation, keeping the learned and enforced components modular. The projection layer is ablated in Sec.~\ref{sec:ablation}.

\subsection{Demonstration Generation}
\label{sec:data}

The model is trained by imitation on style-labeled demonstrations. These are
produced offline by a planner that samples a large set of candidate trajectories
and selects, for each scene and style vector, those minimizing a weighted multi-term
social and goal cost. Retaining several distinct low-cost candidates per scene, rather than the
single optimum, preserves multimodality in the training distribution. The imitated policy generalizes, and in some cases, produces more drastically amplified style differences (Sec.~\ref{sec:ablation}).

The social cost is a sum of norm-grounded terms, each drawn from a documented convention (Table~\ref{tab:costterms}): passing-side preference~\cite{johnson2018passing}, resolution of oncoming and cutting-off encounters through time-to-collision (TTC), following the power-law structure of human avoidance~\cite{karamouzas2014universal}, proxemic
spacing~\cite{hallproxemics, singamaneni2024survey}, and deference to co-moving social groups~\cite{moussaid2010groups}. Instead of assigning each term a free weight, each style axis moves a single interpretable scale. The spatial terms use a varying zone radius that interpolates log-linearly between Hall's intimate ($\sim$0.5 m), personal ($\sim$1.2 m) and social ($\sim$3.7 m) distances \cite{hallproxemics}. Temporal terms incorporate the corresponding distance divided by typical walking speed (1.34 m/s) \cite{SFM} as an anticipation horizon. As a result, the geometry uses scales informed by the proxemics literature rather than being arbitrarily tuned, and at $s_i=-1$, the zone contracts to the intimate radius that every collision-free trajectory already respects. Further cost details appear in the accompanying code.

\begin{table}[t!]
\centering
\caption{Social cost terms and the style axis governing each.}
\label{tab:costterms}
\renewcommand{\arraystretch}{1.1}
\begin{tabular}{@{}lll@{}}
\toprule
Term & Convention captured & Axis \\
\midrule
$g_{\mathrm{prox}}$   & keep clear of pedestrian space         & $s_{\mathrm{prox}}$ \\
$g_{\mathrm{rear}}$   & avoid approaching from directly behind & $s_{\mathrm{prox}}$ \\
$g_{\mathrm{side}}$   & pass on the convention side            & $s_{\mathrm{pass}}$ \\
$g_{\mathrm{ttc}}$    & anticipate and defer to oncomers & $s_{\mathrm{yield}}$ \\
$g_{\mathrm{cut}}$    & do not cut in front of a pedestrian    & $s_{\mathrm{yield}}$ \\
$g_{\mathrm{group}}$    & do not split social groups             & $s_{\mathrm{group}}$ \\
$g_{\mathrm{smooth}}$ & limit jerk and angular acceleration    & --- \\
$g_{\mathrm{goal}}$ & make progress toward the goal    & --- \\
\bottomrule
\end{tabular}
\end{table}

Crucially, each demonstration only carries a non-zero value on one axis. Thus, the model never observes a composed style during training, and composition (Sec.~\ref{sec:sim-results}) is achieved entirely at inference. The scheme is agnostic to the demonstration source, only requiring examples with single-axis style labels. The sampling planner is one convenient way to obtain them at scale. Human-annotated or real-world pedestrian datasets with similar labeling could be substituted. An ablation (Sec.~\ref{sec:ablation}) contrasts performance using joint-labeled demonstrations from the same expert. Scenes use occupancy map crops, exposing the model to realistic indoor geometry.

\section{Evaluation}
\label{sec:eval}

\subsection{Experimental Setup}
\label{sec:eval-setup}

\noindent\textbf{Environment and scenarios.}
Simulation uses the CrowdNav environment~\cite{CrowdNavSARL} with ORCA-driven pedestrians~\cite{ORCA}. Pedestrians are mutually observable and reactive, but do not respond to the robot, requiring the agent to maintain socially compliant behavior without relying on cooperation. Instead of the standard circle-crossing case, episodes come from a more general randomized scenario generator. Here, 500 scenes use random start and goal poses for the robot and $1$--$10$ pedestrians. Open, unmapped scenes are used to accommodate baselines. 

Style evaluation uses a second set of 500 scenes built from geometric scenarios, including corridors, head-on encounters, overtaking, crossing, and stationary or walking groups. These scenarios isolate specific interactions, making behavioral differences between styles more legible than in dense crowds.

\noindent\textbf{Baselines.}
We compare against reactive planners (SFM~\cite{SFM}, ORCA~\cite{ORCA}),
value-, attention-, and graph-based RL policies (CADRL~\cite{CADRL}, LSTM-RL~\cite{LSTM-RL},
SARL~\cite{CrowdNavSARL}, RGL~\cite{RGL}), spatio-temporal graph and transformer
policies (DSRNN~\cite{DSRNN}, NaviSTAR~\cite{NaviSTAR}, HEIGHT~\cite{HEIGHT}),
and an optimization-based planner coupling prediction and planning
(SICNav~\cite{SICNav}). ComposableNav~\cite{ComposableNav} is the closest work conceptually but assumes behavioral instructions absent from our evaluation protocol. Learned baselines are retrained on the same random scenario distribution, with consistent parameters and defaults otherwise. 

Models with a fixed human count are padded, while natively holonomic policies are retrained with \cite{CADRL,LSTM-RL,CrowdNavSARL,RGL} or adapted to \cite{SFM,ORCA,SICNav} unicycle dynamics. In the case of unsuccessful retraining, holonomic dynamics are trained but unicycle limits are enforced during evaluation \cite{DSRNN,NaviSTAR,HEIGHT}. SFM and ORCA use a 0.15~m safety inflation for consistency.

\noindent\textbf{Metrics.}
Standard outcome metrics are success, timeout and collision rate (SR/TR/CR);
time to goal (TTG) and path length (PL) are averaged over successful episodes.
For social conduct, we use metrics informed by the literature~\cite{francis2025}, expressed as rates, and grounded in Hall's
interpersonal zones~\cite{hallproxemics}:

\begin{itemize}
  \item \emph{Mean clearance} (MeanD): the per-step minimum distance to any
  pedestrian, averaged over the episode.
  \item \emph{Personal space intrusion rate} (PSI): the fraction of episode time a pedestrian is in personal space ($\sim$1.2 m).
  \item \emph{Passing-side bias} (PSB): over encounters with moving pedestrians
  inside the social zone ($\sim$3.7 m), the mean signed side on which the robot
  passes at closest approach. Both passing and overtaking on a pedestrian's left match the right-hand ($+$) convention. Values lie in
  $[-1,1]$; $0$ means no consistent side and $\pm1$ a fully consistent one.
  \item \emph{Time-to-collision infraction rate} (TCR): the fraction of episode time
  at which the time to contact with the nearest closing pedestrian falls below a
  1.5~s reaction interval.
  \item \emph{Path cutoff rate} (PCR): the fraction of episode time the robot
  spends inside the forward field of a pedestrian, of length set by
  their speed over the same reaction interval.
  \item \emph{Group split rate} (GSR): the fraction of episode time the robot
  spends inside the zone between a co-moving pair.
\end{itemize}

Each metric is averaged over all episodes, measures the rate of a social event and is independent of the imitated social cost.

\noindent\textbf{Implementation.}
Parameters include $T=32$, $\Delta t=0.05$~s, $P=10$, $K=100$ in training, with DDIM using 20 at inference, and $N=5$. Moreover, $w_s=1$, $\sum_i w_i=5$, $r^r=r^j=0.25$~m, $r^s=0.15$~m, $v_{\max}=1$~m/s, $\omega_{\max}=\pi$~rad/s, $a_{\max}=1.5$~m/s$^2$, $\alpha_{\max}=\pi$~rad/s$^2$ and $5$ SQP iterations are used.

\subsection{Simulation Results}
\label{sec:sim-results}

\begin{table}[t]
\caption{Success, efficiency and proximity metrics for SoGuDiff and baselines on a random set of 500 crowd scenes. For each metric, the best value is bolded and the second-best is underlined. Holonomically trained baselines are denoted by $^\dagger$.}
\label{tab:baseline_comp}
\centering
\begin{tabular}{lccccc}
\toprule
\multirow{2}{*}{\textbf{Method}} & \textbf{SR/TR/CR} & \textbf{TTG} & \textbf{PL} & \textbf{MeanD} & \textbf{PSI} \\
 & $\uparrow$ \textbf{(\%)} & $\downarrow$ \textbf{(s)} & $\downarrow$ \textbf{(m)} & $\uparrow$ \textbf{(m)} & $\downarrow$ \textbf{(\%)} \\
\midrule
SFM \cite{SFM} & 84/\phantom{0}1/15 & 11.12 & \textbf{\phantom{0}9.27} & 3.49 & 11.25 \\
ORCA \cite{ORCA} & \underline{98/\phantom{0}0/\phantom{0}2} & 10.75 & \phantom{0}\underline{9.30} & 3.50 & 11.28 \\
CADRL \cite{CADRL} & 97/\phantom{0}2/\phantom{0}1 & 11.17 & 10.12 & 3.66 & \textbf{\phantom{0}4.78} \\
LSTM-RL \cite{LSTM-RL} & 94/\phantom{0}3/\phantom{0}3 & 11.56 & 10.81 & \underline{3.67} & \phantom{0}\underline{5.20} \\
SARL \cite{CrowdNavSARL} & 96/\phantom{0}2/\phantom{0}2 & 11.40 & 10.32 & \textbf{3.71} & \phantom{0}5.95 \\
RGL \cite{RGL} & 92/\phantom{0}6/\phantom{0}2 & 11.25 & 10.24 & 3.65 & \phantom{0}7.13 \\
DSRNN$^\dagger$ \cite{DSRNN} & 91/\phantom{0}4/\phantom{0}5 & 11.94 & 10.63 & 3.63 & \phantom{0}6.51 \\
NaviSTAR$^\dagger$ \cite{NaviSTAR} & 81/\phantom{0}3/16 & 13.61 & 10.10 & 3.58 & \phantom{0}8.89 \\
SICNav \cite{SICNav} & 96/\phantom{0}0/\phantom{0}4 & \underline{10.31} & \phantom{0}9.67 & 3.44 & 12.56 \\
HEIGHT$^\dagger$ \cite{HEIGHT} & 88/\phantom{0}2/10 & 11.15 & 10.00 & 3.58 & \phantom{0}7.87 \\
SoGuDiff & \textbf{99/\phantom{0}1/\phantom{0}0} & \textbf{10.22} & \phantom{0}9.59 & 3.50 & \phantom{0}6.73 \\

\bottomrule
\end{tabular}
\end{table}

\begin{figure}[t]
    \centering
    \includegraphics[width=0.8\linewidth]{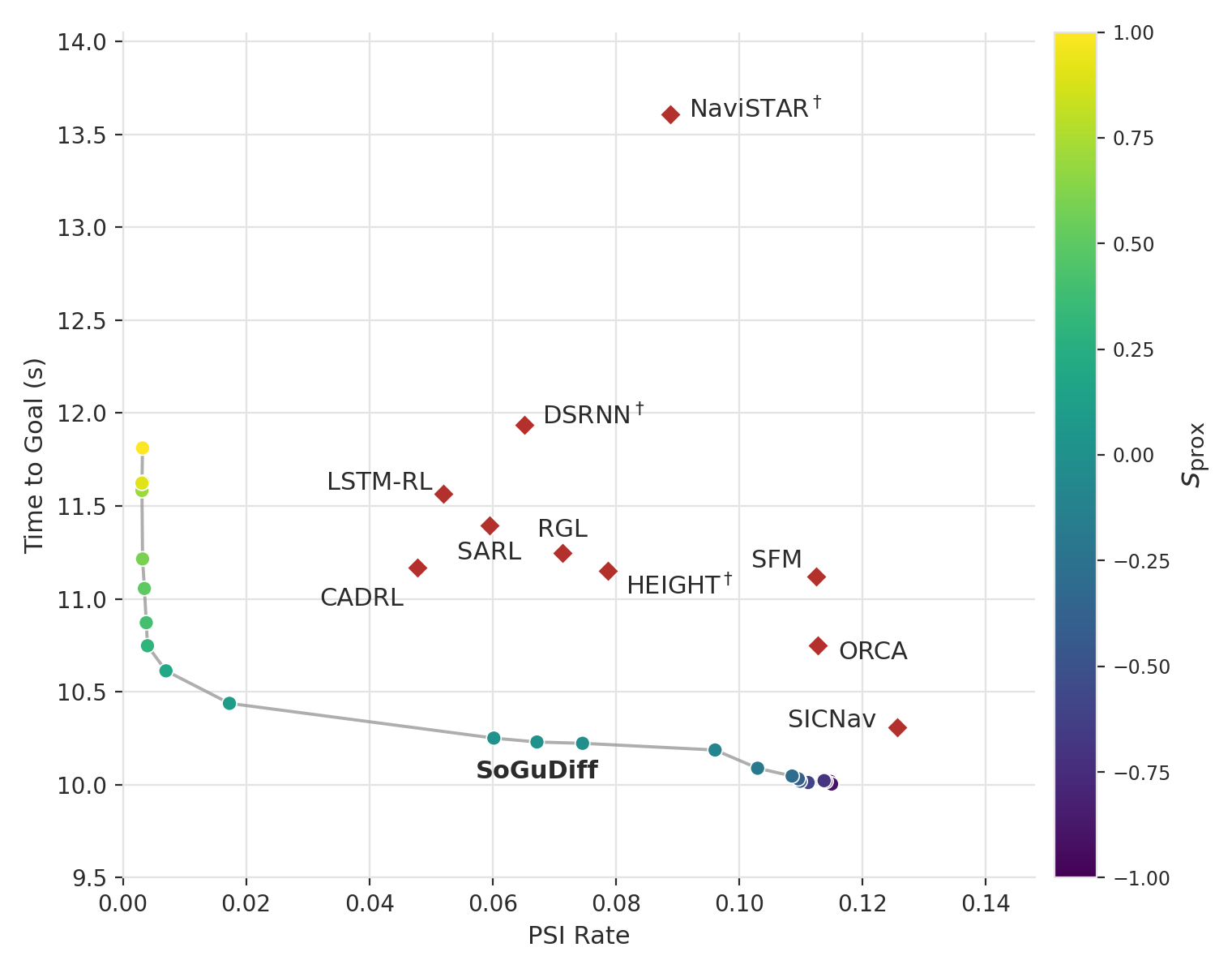}
    \caption{The TTG vs PSI tradeoff for fixed-behavior baselines and SoGuDiff, traced by varying proximity style, over 500 random crowd scenes. Holonomically trained baselines are denoted by $^\dagger$.}
	\label{pareto}
\end{figure}

\begin{figure*}[t!]
\centering
\begin{subfigure}[b]{0.245\textwidth}
    \centering
    \includegraphics[width=\textwidth]{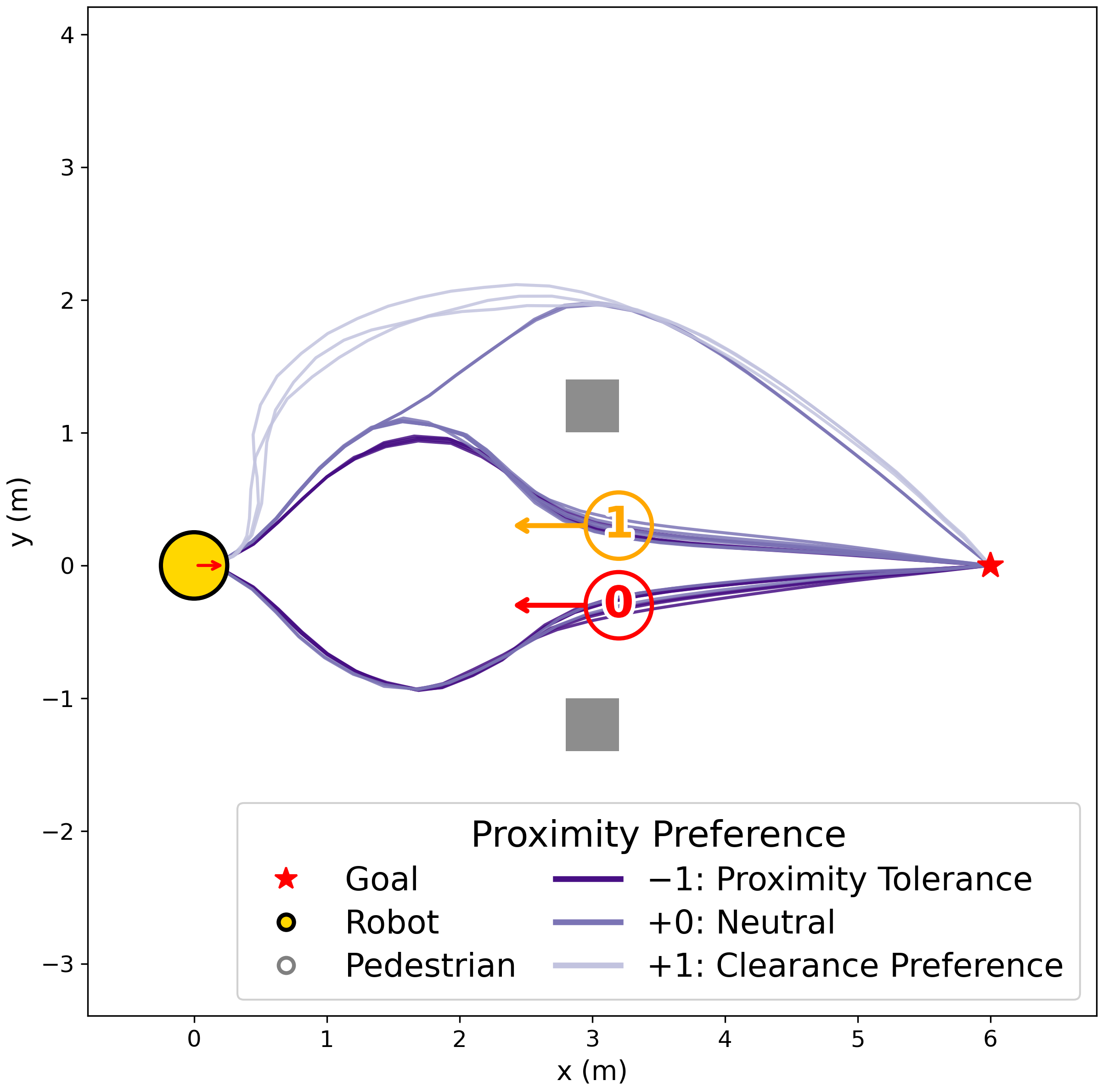}
    \caption{Proximity Preference}
    \label{fig:sub1}
\end{subfigure}
\hfill
\begin{subfigure}[b]{0.245\textwidth}
    \centering
    \includegraphics[width=\textwidth]{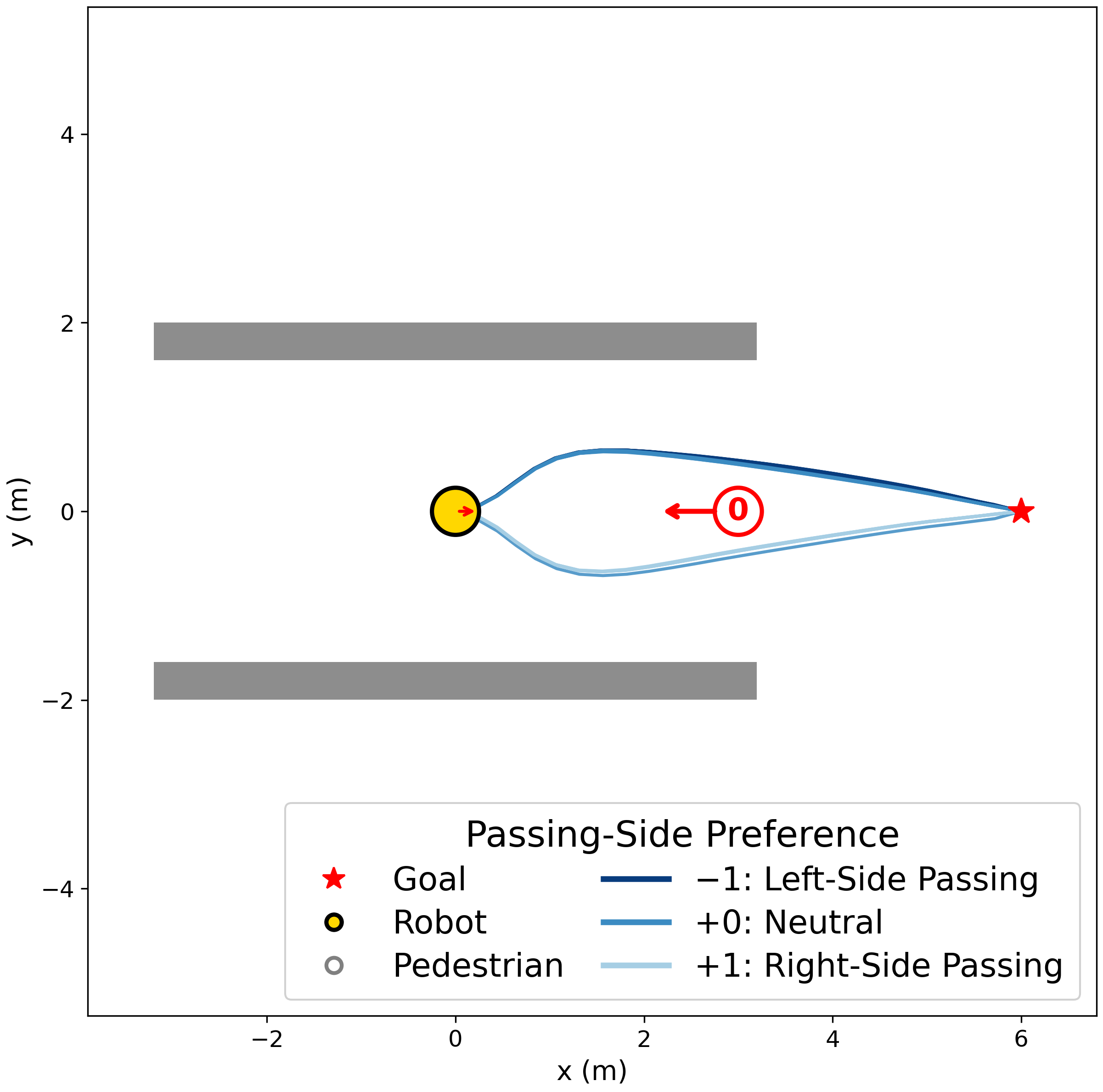}
    \caption{Passing-Side Preference}
    \label{fig:sub2}
\end{subfigure}
\hfill
\begin{subfigure}[b]{0.245\textwidth}
    \centering
    \includegraphics[width=\textwidth]{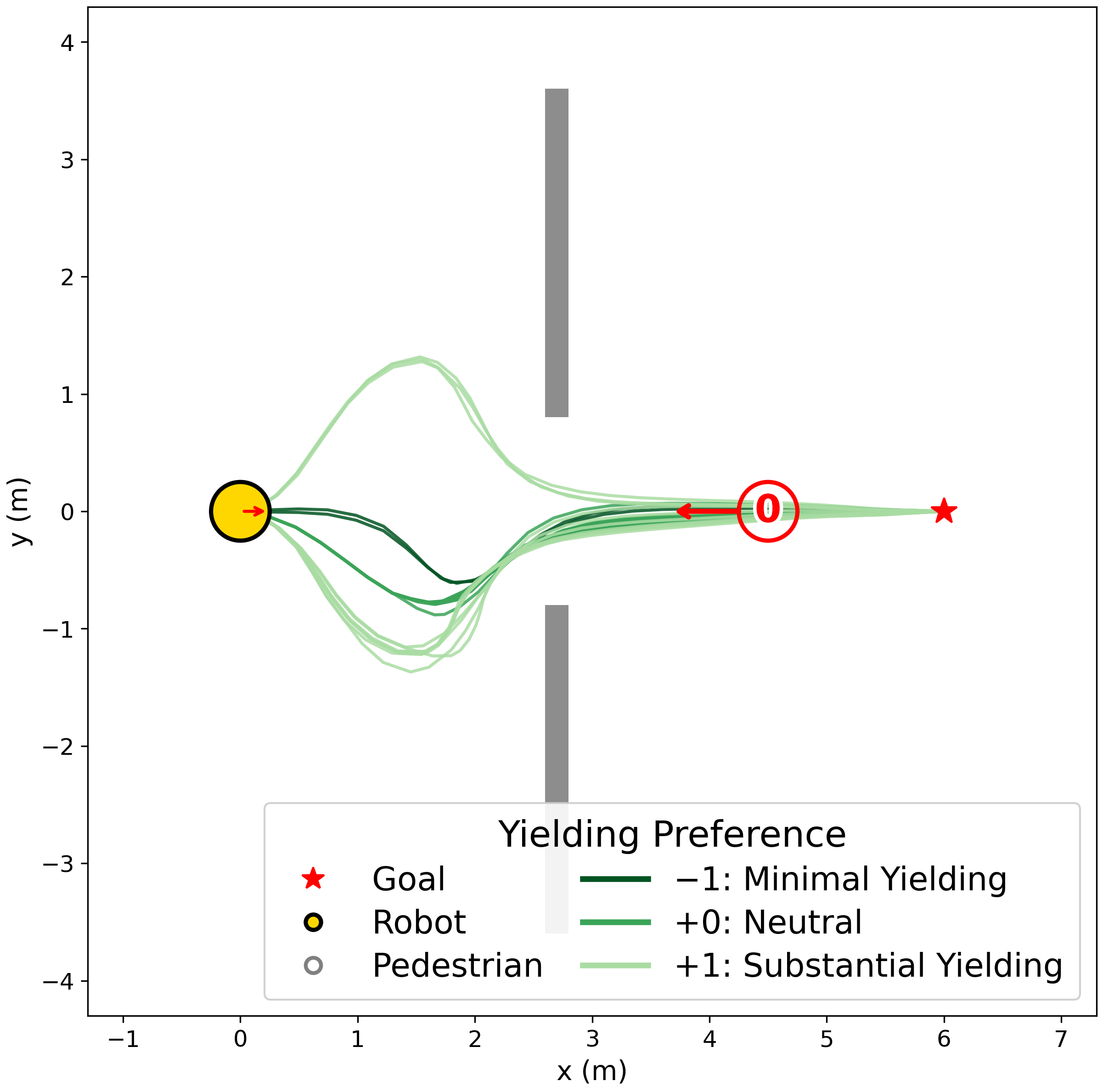}
    \caption{Yielding Preference}
    \label{fig:sub3}
\end{subfigure}
\hfill
\begin{subfigure}[b]{0.245\textwidth}
    \centering
    \includegraphics[width=\textwidth]{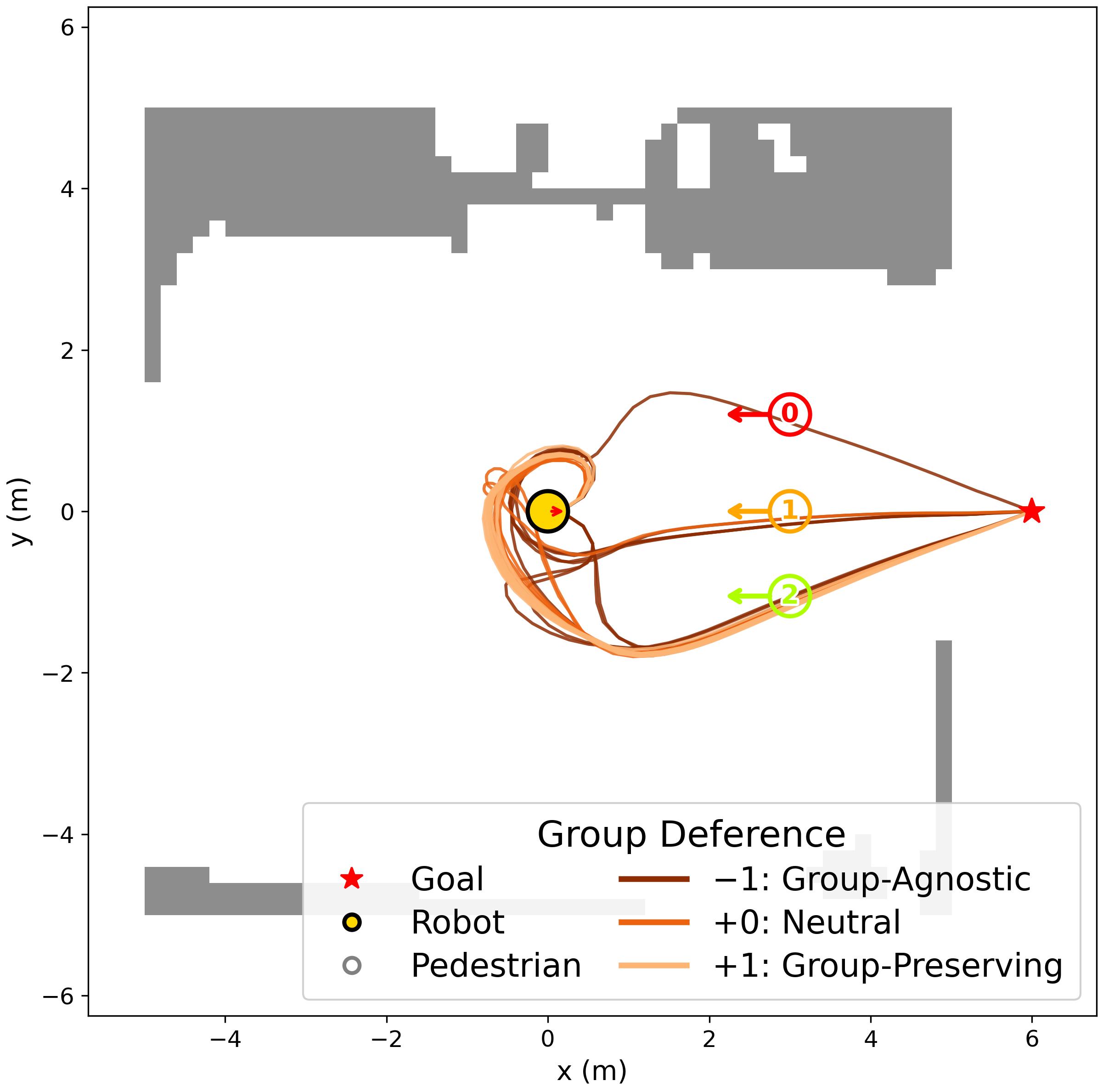}
    \caption{Group Deference}
    \label{fig:sub4}
\end{subfigure}
\caption{Closed-loop trajectory distributions for the nondeterministic SoGuDiff planner using swept single-axis styles across 10 randomly seeded runs on representative scenes. Darker trajectories correspond to $-1$ styles, with lighter corresponding to $+1$ for the (a) proximity, (b) passing-side, (c) yielding and (d) group deference style axis sweeps.}
\label{fig:sweep}
\end{figure*}

\noindent\textbf{Baseline comparison.}
SoGuDiff performance using the neutral style $\mathbf{s}=\mathbf{0}$ is reported against the baselines in Table~\ref{tab:baseline_comp}. Our approach attains the highest success rate, the lowest time to goal, and remains otherwise competitive with the state-of-the-art. Notably, our policy spans a range of behaviors while the baselines represent a single point. At the neutral style, proxemic behavior sits between the reactive and optimization-based planners (which achieve short paths but frequently intrude personal space) and the RL policies (which maintain distance but take longer). 

Fig.~\ref{pareto} sweeps the proxemic axis alone in increments from $-1$ to $+1$, illustrating a frontier that strictly dominates baselines along the efficiency-vs-sociality tradeoff. This strong, runtime-configurable behavior accommodates varying social contexts, while other style axes exhibit similar tradeoffs.

\begin{table}[t!]
\newcommand{\ps}[1]{\makebox[0.6em][r]{#1}}
\caption{SoGuDiff behavior on a set of 500 geometric scenarios, using swept single-axis styles. For each swept axis, values pertaining to the style-specific metrics are shaded.}
\label{tab:singlesweep}
\centering
\begin{tabular}{l@{\hskip 8.8pt}c@{\hskip 8.8pt}c@{\hskip 8.8pt}c@{\hskip 8.8pt}c@{\hskip 8.8pt}c@{\hskip 8.8pt}c}
\toprule
\multirow{2}{*}{\textbf{Social Style}} & \textbf{SR/TR/CR} & \textbf{MeanD} & \textbf{PSB} & \textbf{TCR} & \textbf{PCR} & \textbf{GSR} \\
 & $\uparrow$ \textbf{(\%)} & $\uparrow$ \textbf{(m)} & $\updownarrow$ \textbf{(-)} & $\downarrow$ \textbf{(\%)} & $\downarrow$ \textbf{(\%)} & $\downarrow$ \textbf{(\%)} \\
\midrule
Neutral & \phantom{0}96/\phantom{0}4/\phantom{0}0 & 2.38 & \ps{+}0.071 & 0.38 & 2.31 & 17.97 \\
\midrule
prox\ps{+}1 & \phantom{0}95/\phantom{0}5/\phantom{0}0 & \cellcolor{gray!20} 3.28 & \ps{+}0.232 & 0.02 & 0.00 & \phantom{0}0.00 \\
prox\ps{+}0.5 & \phantom{0}91/\phantom{0}9/\phantom{0}0 & \cellcolor{gray!20} 2.75 & \ps{+}0.181 & 0.03 & 0.01 & \phantom{0}0.55 \\
prox\ps{-}0.5 & 100/\phantom{0}0/\phantom{0}0 & \cellcolor{gray!20} 2.29 & \ps{-}0.111 & 1.94 & 2.93 & 21.12 \\
prox\ps{-}1 & 100/\phantom{0}0/\phantom{0}0 & \cellcolor{gray!20} 2.27 & \ps{-}0.148 & 2.37 & 3.40 & 20.32 \\
\midrule
pass\ps{+}1 & 100/\phantom{0}0/\phantom{0}0 & 2.59 & \cellcolor{gray!20}\ps{+}0.763 & 0.53 & 1.10 & 8.83 \\
pass\ps{+}0.5 & 100/\phantom{0}0/\phantom{0}0 & 2.50 & \cellcolor{gray!20}\ps{+}0.744 & 0.52 & 1.15 & 11.21 \\
pass\ps{-}0.5 & \phantom{0}99/\phantom{0}1/\phantom{0}0 & 2.49 & \cellcolor{gray!20}\ps{-}0.681 & 0.52 & 1.53 & 11.28 \\
pass\ps{-}1 & 100/\phantom{0}0/\phantom{0}0 & 2.55 & \cellcolor{gray!20}\ps{-}0.705 & 0.54 & 1.35 & \phantom{0}9.08 \\
\midrule
yield\ps{+}1 & \phantom{0}97/\phantom{0}3/\phantom{0}0 & 2.44 & \ps{-}0.062 & \cellcolor{gray!20}0.19 & \cellcolor{gray!20}0.58 & 10.62 \\
yield\ps{+}0.5 & \phantom{0}95/\phantom{0}5/\phantom{0}0 & 2.39 & \ps{+}0.075 & \cellcolor{gray!20}0.31 & \cellcolor{gray!20}1.14 & 15.63 \\
yield\ps{-}0.5 & \phantom{0}96/\phantom{0}4/\phantom{0}0 & 2.37 & \ps{+}0.158 & \cellcolor{gray!20}0.65 & \cellcolor{gray!20}2.69 & 20.64 \\
yield\ps{-}1 & \phantom{0}96/\phantom{0}4/\phantom{0}0 & 2.36 & \ps{+}0.162 & \cellcolor{gray!20}1.03 & \cellcolor{gray!20}3.00 & 20.38 \\
\midrule
group\ps{+}1 & 100/\phantom{0}0/\phantom{0}0 & 2.42 & \ps{+}0.017 & 0.34 & 1.18 & \cellcolor{gray!20}\phantom{0}7.54 \\
group\ps{+}0.5 & \phantom{0}97/\phantom{0}3/\phantom{0}0 & 2.40 & \ps{+}0.073 & 0.40 & 1.58 & \cellcolor{gray!20}13.33 \\
group\ps{-}0.5 & \phantom{0}95/\phantom{0}5/\phantom{0}0 & 2.37 & \ps{+}0.101 & 0.40 & 2.33 & \cellcolor{gray!20}19.62 \\
group\ps{-}1 & \phantom{0}95/\phantom{0}5/\phantom{0}0 & 2.37 & \ps{+}0.119 & 0.36 & 2.57 & \cellcolor{gray!20}20.00 \\

\bottomrule
\end{tabular}
\end{table}

\noindent\textbf{Single-axis sweeps.}
In Table~\ref{tab:singlesweep}, each axis is swept over $\{-1,-0.5,0,+0.5,+1\}$ with the remaining axes held neutral. Each axis moves its associated metric monotonically and substantially.
Proxemic conservatism raises mean clearance from 2.27~m at $-1$ to 3.28~m at
$+1$, with a neutral value of 2.38~m. Passing-side bias moves steeply from right-side dominant where scene geometry allows ($+0.763$) to left-side dominant ($-0.705$), while neutral carries no preference ($+0.071$). The yield axis sweep reduces the TTC infraction rate from 1.03\% to 0.19\% and path cutoff rate from 3.00\% to 0.58\%. Group deference reduces the group split rate from 20.00\% to 7.54\%.
Timeout rates remain below 10\% even for conservative styles, and no style produces a collision.

Evidently, behavioral couplings exist across the distinct axes due to shared geometry. Increased proxemic conservatism improves yielding and group adherence metrics. Additionally, increased side-preference and yielding both increase group adherence and mean clearance. These secondary effects illustrate that the styles share commitment to sociality. Fig.~\ref{fig:sweep} shows the trajectory distributions for varying single-axis styles in four representative scenarios. Visually, styles attain distinct modes that match the behavior they are expected to induce.

\begin{figure}[t!]
    \centering
    \includegraphics[width=0.75\linewidth]{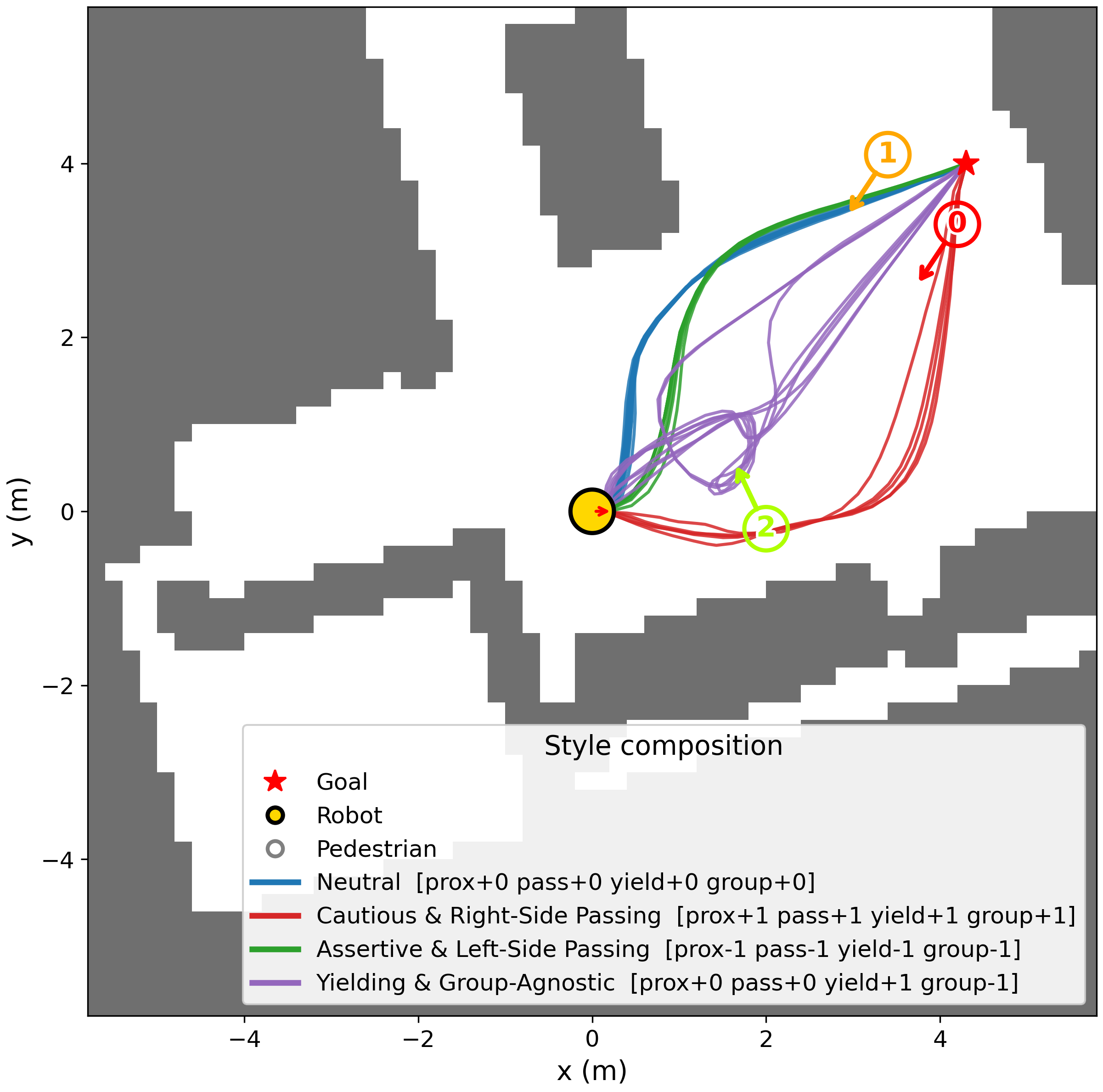}
    \caption{Closed-loop trajectory distributions for the SoGuDiff planner using composed axis styles across 10 randomly seeded runs each.}
	\label{fig:compose}
\end{figure}

\noindent\textbf{Style composition.}
Distinct behaviors are visualized for several composed styles in Fig.~\ref{fig:compose}, while composition is evaluated quantitatively in Table~\ref{tab:composition}. Despite training on only single-axis demonstrations, the guidance scheme achieves composed behavior. With all four axes set to $+1$, mean clearance rises
to 2.87~m, cutoffs fall to 0.18\% and group splits to 0.97\%, while right-side passing is preferred ($+0.405$). Setting all axes to $-1$ moves every metric in the direction of reduced sociality and left-side passing. The $[0,+1,+1,0]$ style induces right-side passing and yielding behaviors, whereas $[0,-1,+1,-1]$ exhibits left-side passing and greater group agnosticism. 

Notably, composition is not linear. Certain axes tend to dominate in conflicting styles, such as increased assertiveness over group deference in the $[-1,0,0,+1]$ style. Axis adherence is generally attenuated
relative to the corresponding single-axis style, most visibly for the passing axis at the $[+1]^4$ and $[-1]^4$ styles. This plausibly reflects the reduced per-axis weight under a fixed guidance budget. Nonetheless, composed styles generally produce behavior consistent with each composed axis, while remaining interpretable even in conflicting configurations.

\begin{table}[t!]
\newcommand{\sv}[1]{\makebox[1.1em][r]{#1}} 
\newcommand{\ps}[1]{\makebox[0.6em][r]{#1}}
\caption{SoGuDiff behavior on a set of 500 geometric scenarios, using composed axis styles. Social styles follow the $[s_{\mathrm{prox}},s_{\mathrm{pass}},s_{\mathrm{yield}},s_{\mathrm{group}}]$ notation.}
\label{tab:composition}
\centering
\begin{tabular}{c@{\hskip 7.9pt}c@{\hskip 7.9pt}c@{\hskip 7.9pt}c@{\hskip 7.9pt}c@{\hskip 7.9pt}c@{\hskip 7.9pt}c}
\toprule
\multirow{2}{*}{\textbf{Social Style}} & \textbf{SR/TR/CR} & \textbf{MeanD} & \textbf{PSB} & \textbf{TCR} & \textbf{PCR} & \textbf{GSR} \\
 & $\uparrow$ \textbf{(\%)} & $\uparrow$ \textbf{(m)} & $\updownarrow$ \textbf{(-)} & $\downarrow$ \textbf{(\%)} & $\downarrow$ \textbf{(\%)} & $\downarrow$ \textbf{(\%)} \\
\midrule
$[$\sv{-1},\sv{0},\sv{0},\sv{+1}$]$   & 100/\phantom{0}0/\phantom{0}0 & 2.29 & \ps{-}0.081 & 2.00 & 2.79 & 17.23 \\
$[$\sv{0},\sv{+1},\sv{+1},\sv{0}$]$   & 100/\phantom{0}0/\phantom{0}0 & 2.48 & \ps{+}0.647 & 0.27 & 0.61 & \phantom{0}8.94 \\
$[$\sv{+1},\sv{-1},\sv{-1},\sv{0}$]$  & 100/\phantom{0}0/\phantom{0}0 & 2.49 & \ps{-}0.249 & 0.57 & 1.46 & \phantom{0}7.49 \\
$[$\sv{0},\sv{-1},\sv{+1},\sv{-1}$]$  & \phantom{0}98/\phantom{0}2/\phantom{0}0 & 2.45 & \ps{-}0.412 & 0.40 & 1.01 & 11.37 \\
$[$\sv{+1},\sv{+1},\sv{+1},\sv{+1}$]$ & 100/\phantom{0}0/\phantom{0}0 & 2.87 & \ps{+}0.405 & 0.06 & 0.18 & \phantom{0}0.97 \\
$[$\sv{-1},\sv{-1},\sv{-1},\sv{-1}$]$ & \phantom{0}99/\phantom{0}1/\phantom{0}0 & 2.39 & \ps{-}0.395 & 0.82 & 2.06 & 17.42 \\
\bottomrule
\end{tabular}
\end{table}

\subsection{Ablations}
\label{sec:ablation}

Components of the method are isolated in Table~\ref{tab:ablations}.

\noindent\textbf{Per-axis vs joint composition.}
In contrast to the per-axis guidance scheme in \eqref{eq:cfg}, joint guidance uses a single joint conditional pass for all styles. Maintaining the single-axis training scheme and trained on every conditioning mask, joint guidance fails to coherently compose passing-side bias for the $[-1]^4$ style. Meanwhile, when joint-axis demonstrations are used in training, joint guidance replicates the expert accurately, while per-axis does not. Thus, matching the guidance scheme with the demonstration source format produces the best results. The single-axis scheme generalizes, approaching joint composition despite needing only single-axis-labeled demonstrations, and permitting a unique guidance weight per axis.

\noindent\textbf{Guidance weight.}
Sweeping the total guidance weight over $\{1,3,10\}$ with the style $s_{\mathrm{yield}}=+1$ strengthens adherence up to saturation.
Higher weights also compress the distinction between intermediate and extreme style values, as both saturate toward the same guided extreme. As trajectories are pushed further off the training manifold, success rate slightly declines. Hence, styles $\mathbf{s}^{(i)}$ and weights $w_i$ are complementary controls.

\noindent\textbf{Online expert.}
The full-sample-budget demonstration sampler is run online as a policy, with an ensuing projection layer. For the $s_{\mathrm{prox}}$ and $s_{\mathrm{pass}}=-1$ styles, the imitated planner (using CFG extrapolation) achieves more drastic stylistic differences. The expert is more adherent to groups, causing timeouts, where the imitated policy carries the challenge of characterizing and avoiding groups implicitly. We find that in general, the imitated policy matches, and in some cases exceeds, the expert, while remaining agnostic to demonstration source.

\noindent\textbf{Feasibility projection.}
Removing the projection layer and directly clamping the diffused path to feasible control actions reduces successes, raising timeouts and collisions. Mean clearance rises to 4.08~m, reflecting failure, not caution, as the difference between planning and action causes stalled motion.

\begin{table}[t!]
\newcommand{\sv}[1]{\makebox[1.1em][r]{#1}} 
\newcommand{\ps}[1]{\makebox[0.6em][r]{#1}}
\caption{Ablation results for composition scheme, guidance weight, online expert policy and feasibility projection layer on a set of 500 geometric scenarios.}
\label{tab:ablations}
\centering
\begin{tabular}{l@{\hskip 0.7pt}c@{\hskip 2.9pt}c@{\hskip 2.5pt}c@{\hskip 3.5pt}c@{\hskip 2.9pt}c@{\hskip 2.9pt}c}
\toprule
\multirow{2}{*}{\textbf{Style, Ablation}} & \textbf{SR/TR/CR} & \textbf{MeanD} & \textbf{PSB} & \textbf{TCR} & \textbf{PCR} & \textbf{GSR} \\
 & $\uparrow$ \textbf{(\%)} & $\uparrow$ \textbf{(m)} & $\updownarrow$ \textbf{(-)} & $\downarrow$ \textbf{(\%)} & $\downarrow$ \textbf{(\%)} & $\downarrow$ \textbf{(\%)} \\
\midrule
\makecell[l]{$[$\sv{-1},\sv{-1},\sv{-1},\sv{-1}$]$, sing.-ax. \\train, joint guid.} & 100/\phantom{0}0/\phantom{0}0 & 2.28 & \ps{+}0.026 & 1.95 & 2.77 & 20.18 \\
\makecell[l]{group\ps{+}1, joint train,\\joint guid.} & 100/\phantom{0}0/\phantom{0}0 & 2.43 & \ps{+}0.078 & 0.50 & 1.34 & \phantom{0}5.46 \\
\makecell[l]{group\ps{+}1, joint train, \\per-ax. guid.} & \phantom{0}95/\phantom{0}5/\phantom{0}0 & 2.33 & \ps{+}0.193 & 0.40 & 2.96 & 19.35 \\
\midrule
yield\ps{+}1, $\sum_i w_i=1$   & \phantom{0}97/\phantom{0}3/\phantom{0}0 & 2.42 & \ps{+}0.046 & 0.29 & 1.00 & 14.64 \\
yield\ps{+}1, $\sum_i w_i=3$   & \phantom{0}96/\phantom{0}4/\phantom{0}0 & 2.42 & \ps{-}0.001 & 0.23 & 0.66 & 11.72 \\
yield\ps{+}1, $\sum_i w_i=10$   & \phantom{0}95/\phantom{0}5/\phantom{0}0 & 2.41 & \ps{-}0.035 & 0.24 & 0.56 & 12.34 \\
\midrule
prox\ps{-}1, online expert   & \phantom{0}96/\phantom{0}2/\phantom{0}2 & 2.35 & \ps{+}0.117 & 0.34 & 2.61 & 20.15 \\
pass\ps{-}1, online expert   & \phantom{0}97/\phantom{0}3/\phantom{0}0 & 2.51 & \ps{-}0.567 & 0.45 & 1.78 & 14.45 \\
group\ps{+}1, online expert  & \phantom{0}74/26/\phantom{0}0 & 2.24 & \ps{+}0.101 & 0.23 & 3.49 & \phantom{0}1.02 \\
\midrule
Neutral style, no proj.   & \phantom{0}70/26/\phantom{0}4 & 4.08 & \ps{+}0.105 & 3.76 & 2.00 & \phantom{0}9.61 \\
\bottomrule
\end{tabular}
\end{table}

\subsection{Real-World Deployment}
\label{sec:eval-real}

\begin{figure}[t!]
    \centering
    \includegraphics[width=1.0\linewidth]{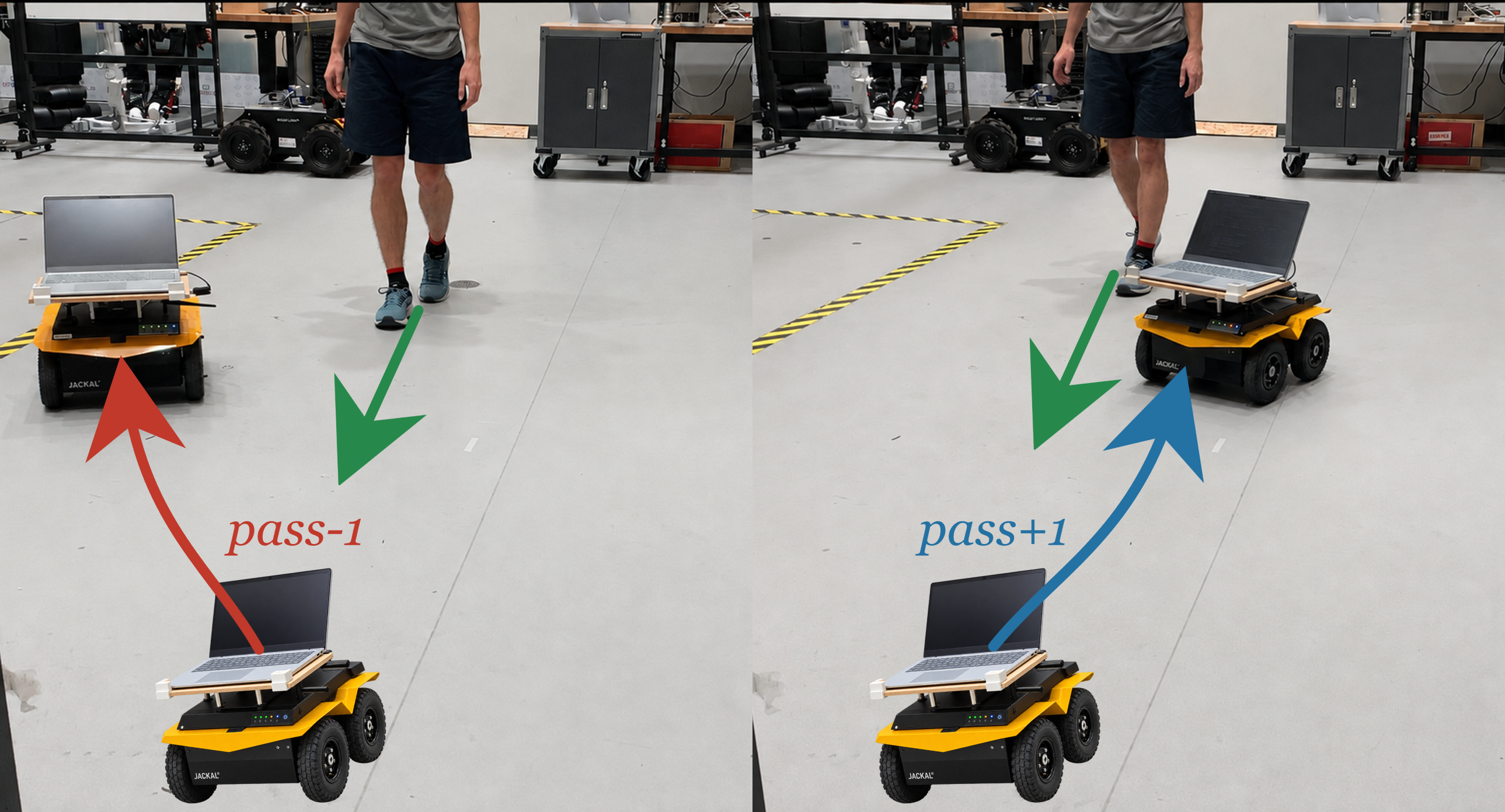}
    \caption{Physical demonstration results for the distinct left-side passing (left) and right-side passing (right) styles on a repeated test.}
	\label{DemoFig}
\end{figure}

We deploy SoGuDiff on a Clearpath Jackal with pedestrian detection (via YOLO11n) and tracking, running the full pipeline at 10 Hz using a tethered laptop with an AMD Ryzen 7 7745HX CPU and NVIDIA RTX 4070 GPU. Collision-free, distinctly styled behavior is reproduced on physical hardware (Fig.~\ref{DemoFig}). Dedicated scenes contrast $\pm1$ per-axis and composed styles across repeated runs, while additional scenes show neutral and runtime-varying styles. Trials are video-recorded\footnote{Video is available at \url{https://youtu.be/wsqTSdGBZrw}.}.

\section{Discussion}
\label{sec:discussion}

\noindent\textbf{Limitations.}
Despite the strengths of our approach, we note some limitations. The projection layer makes assumptions and offers no safety guarantees. Pedestrians are propagated at a constant estimated velocity, solver convergence is not always achievable, and the controlled stop fallback reduces but does not eliminate collision severity. Overly proxemic conservative styles can induce freezing in dense crowds. Guidance promotes a style with higher likelihood rather than guaranteeing it. In conflicting composed styles, a single style axis may dominate behavior, and excessive guidance weights degrade continuous stylistic control. Finally, groups are only characterized implicitly, making group deference more difficult to learn.

\noindent\textbf{Conclusion.}
This paper introduced SoGuDiff, a diffusion planner whose social conduct is specified at deployment along four interpretable axes rather than fixed during training. Trained using single-axis demonstrations, styles are composed at inference through per-axis CFG. A feasibility projection keeps kinematic and clearance concerns separate from the learned social behavior. SoGuDiff covers an interpretable, continuous span of social behaviors and is shown to strictly dominate fixed-behavior baselines over an efficiency-vs-sociality tradeoff. A natural extension is adaptive style selection during runtime, based on the perceived social setting.

\bibliographystyle{IEEEtran}
\bibliography{references}

\end{document}